\documentclass[sigconf]{acmart}

\AtBeginDocument{%
  \providecommand\BibTeX{{%
    \normalfont B\kern-0.5em{\scshape i\kern-0.25em b}\kern-0.8em\TeX}}}

\setcopyright{acmlicensed}
\copyrightyear{2018}
\acmYear{2018}
\acmDOI{XXXXXXX.XXXXXXX}

\acmConference[Conference acronym 'XX]{Make sure to enter the correct
  conference title from your rights confirmation emai}{June 03--05,
  2018}{Woodstock, NY}
\acmISBN{978-1-4503-XXXX-X/18/06}

\usepackage{enumitem}
\usepackage{multirow}
\usepackage{graphicx}
\usepackage{booktabs}
\usepackage{graphicx}
\usepackage{adjustbox}
\usepackage{cleveref}
\usepackage{colortbl}
\makeatletter

\newcommand{\Rmnum}[1]{\expandafter\@slowromancap\romannumeral #1@}
\makeatother

\begin{document}

\title{MMA-UNet: A Multi-Modal Asymmetric UNet Architecture for Infrared and Visible Image Fusion}

\author{Jingxue Huang}
\authornote{Both authors contributed equally to this research.}
\email{jasonwong30@163.com}
\affiliation{%
  \institution{Foshan University}
  \country{China}
}

\author{Xilai Li}
\authornotemark[1]
\email{20210300236@stu.fosu.edu.cn}
\affiliation{%
  \institution{Foshan University}
  \country{China}
}

\author{Tianshu Tan}
\email{ttanad@connect.ust.hk}
\affiliation{%
  \institution{Hong Kong University of Science and Technology}
  \country{China}
}

\author{Xiaosong Li}
\authornote{Corresponding author.}
\email{lixiaosong@buaa.edu.cn}
\affiliation{%
  \institution{Foshan University}
  \country{China}
}

\author{Tao Ye}
\email{ayetao198715@163.com}
\affiliation{%
  \institution{China University of Mining and Technology (Beijing)}
  \country{China}
}



\renewcommand{\shortauthors}{Trovato and Tobin, et al.}
\renewcommand\footnotetextcopyrightpermission[1]{}
\settopmatter{printacmref=false} 

\begin{abstract}
Multi-modal image fusion (MMIF) maps useful information from various modalities into the same representation space, thereby producing an informative fused image. However, the existing fusion algorithms tend to symmetrically fuse the multi-modal images, causing the loss of shallow information or bias towards a single modality in certain regions of the fusion results. In this study, we analyzed the spatial distribution differences of information in different modalities and proved that encoding features within the same network is not conducive to achieving simultaneous deep feature space alignment for multi-modal images. To overcome this issue, a Multi-Modal Asymmetric UNet (MMA-UNet) was proposed. We separately trained specialized feature encoders for different modal and implemented a cross-scale fusion strategy to maintain the features from different modalities within the same representation space, ensuring a balanced information fusion process. Furthermore, extensive fusion and downstream task experiments were conducted to demonstrate the efficiency of MMA-UNet in fusing infrared and visible image information, producing visually natural and semantically rich fusion results. Its performance surpasses that of the state-of-the-art comparison fusion methods.
\end{abstract}




\keywords{Multi-modal architecture, asymmetric unet, infrared and visible image fusion.}



\maketitle

\section{Introduction}
Infrared and visible image fusion (IVIF) technique integrates useful information captured by different modal sensors to present a comprehensive interpretation of a target scene \cite{r1,r2,r3,r4,r6,r51}. A visible image (VI) effectively captures the global details and color information of the target scene, whereas an infrared image (IR) excels at highlighting the temperature information. A comprehensive perception of various lighting conditions and complex environments can be achieved by effectively fusing the information from these two modalities. Furthermore, IVIF can effectively assist to downstream tasks such as object detection \cite{r7,r8}, semantic segmentation \cite{r9,r10}, and depth estimation \cite{r11,r12}, among others.

In recent years, IVIF has been broadly classified into two main categories: traditional image processing methods that extract features from multi-modal images \cite{r13,r14,r15,r16}, and neural network architectures that train high-performance image fusion models \cite{r17,r18,r19,r20,r21,r22,r23}. In pursuit of stronger generalization capabilities and fusion accuracy, most recent efforts are focused on neural network architectures. IVIF is an atypical image inverse problem in the field of image processing, not guided by ground truth. Therefore, the effective supervised learning paradigm in deep learning is difficult to introduced to IVIF. Whereas the existing unsupervised IVIF methods can achieve an attractive fusion result. However, fusion in the presence of inconsistent information space distributions among multi-modal images remains an unresolved problem. Due to the potential problem, the fusion of different modal images within a single framework may cause loss of source image information or bias towards the features of a single modality. For example, Zhao et al. \cite{r3} proposed a dual-branch feature decomposition IVIF that utilizes Transformer and convolution layer to extract the global and local features of the images, respectively. However, the feature decomposition pattern only enables the network to learn notable details or energy from multi-modal images without considering the feature interaction between them, producing a non-uniform distribution of multi-modal information in the fusion results. Li et al. \cite{r5} introduced graph interaction modules to facilitate the interaction learning of cross-modality features. The shallow features from different modal images must be extracted before feature interaction. However, due to ignore the inconsistent information space distribution between multi-modal image, it cannot be ensured that the multi-modal features inputted into the graph interaction modules during interaction learning are located within the same representation space. This results in a decrease in the performance of interaction learning strategies. Additionally, some researchers \cite{r24} introduced attention mechanisms to achieve cross-modality perception during the feature extraction process. This approach involves separately generating attention maps for the features of each modality and then cross-guiding the reconstruction of the multi-modal features. However, interactive guidance may not necessarily assist in effective feature extraction. When the information content of one modal image is small, the generated attention maps may mislead the network when identifying useful features. Luo et al. \cite{r25} proposed a fusion method based on separate representation learning to achieve a disentangled representation of images and distinguish common and private features among multi-modal images. They followed the principle that multi-modal images only share the same common features and used the formula, "private features + common features = source image," to constrain the network. However, owing to the inherent differences in the same scene captured by different modalities, the common features between the images may not be exactly equal. They represent similar aspects of the same thing, such as details in the VI and faint textures in the IR. In summary, the aforementioned methods, whether from the perspective of global-local, cross-modal feature interaction, cross-attention guidance, or common-private features, do not consider the information space distributions differences between the multi-modal features. Therefore, complex or fixed paradigms are required to infer superior fusion performance.

\begin{figure}[h]
   \centering
   \includegraphics[width=0.98\linewidth]{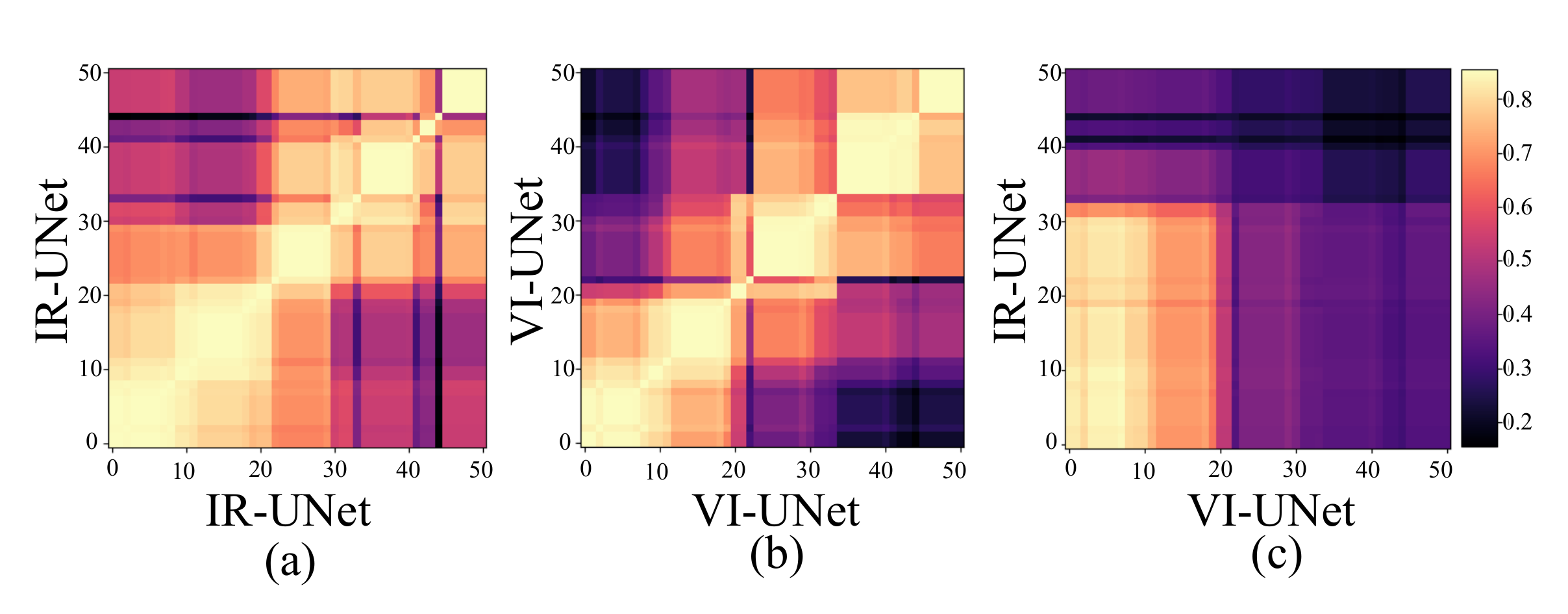}
   \caption{Centered Kernel Alignment. (a) and (b) compute the CKA similarity between all pairs of layers in a single neural network. (c) computes the CKA similarity between all pairs of layers of IR-UNet and VI-UNet. The x and y axes represent the indexing layers.}
   \label{fig1}
\end{figure}

In this study, we reconsidered the approach to multi-modal feature extraction for IVIF and designed a simple and effective asymmetric fusion structure to overcome the problem of inconsistent information space distribution in multi-modal image. We analyzed the differences in the spatial distribution of the multi-modal features and observed \textbf{variations in the speed at which the same network reaches the deep semantic space in different modal images}. The main contributions are summarized as follows:

\begin{itemize}
\item We proposed a multi-modal asymmetric UNet (MMA-UNet) architecture for IVIF that uses a simple and effective method for fusing multi-modal features when compared with the existing complex paradigms.
    
\item We discovered the spatial information distribution differences between IR and VI and derived an asymmetric network. The proposed manner effectively retains the different modalities features in the same representation space, presenting a new fusion paradigm for MMIF.

\item We verified the effectiveness and superiority of MMA-UNet on existing public datasets, where it outperformed the existing the state-of-the-art algorithms on downstream tasks.
\end{itemize}
\textbf{Motivation.}
The existing IVIF algorithms \cite{r17,r18,r19,r20,r21,r55} primarily focus on effectively learning the multi-modal feature interactions while ignoring the inconsistency of the spatial information distribution in multi-modal image. Therefore, we rethought the multi-modal fusion paradigm. Firstly, the two unique UNets were trained separately on IR and VI dataset, represented by IR-UNet and VI-UNet. We then calculated the Centered Kernel Alignment (CKA) \cite{r26} similarity of the features extracted by the encoders of the two UNets. As shown in \cref{fig1}(a) and (b), note that the total number of layers is much greater than the stated depth of the UNet. The latter only accounts for the convolutional layers in the network, but we include all intermediate representations. Specifically, for the UNet framework, the first convolution block contains 7 layers of feature intermediate representation, and the second to fifth convolution blocks contain 11 layers of feature intermediate representation. Neural networks tend to initially learn shallow features, and as the network depth increases, they gradually dive into the deeper semantic spaces for feature learning. The shallow features are mostly similar. However, as the network depth increases, the differences between the deep and shallow features increase. IR captures shallow features in the first 22 layers since they share a greater similarity with the surrounding features, as shown in \cref{fig1}(a). When the network depth exceeds 22 layers, the extracted features vary significantly from the surroundings, manifested as deeper colors in the \cref{fig1}. Conversely, the feature extraction network of the VI begins to show its uniqueness from the 12th layer, and the difference in the subsequent features continues to increase. \textbf{Therefore, we observed that VI can reach deeper semantic spaces faster than IR under the same architecture, and VI can extract shallow information 10 layers faster than IR in our network.}

Several studies \cite{r3,r25} have reported that multi-modal images can be used to distinguish between public and private features in MMIF. Public features share high similarity, whereas the private features do not. As shown in \cref{fig1}(c), we computed the CKA similarity between the features of each layer of VI and IR.  The first 30 layers of the IR features are highly similar to the first 20 layers of the VI features, showed that the extraction of IR private features is slower than that of VI. Additionally, this verifies the above conclusion again VI reaches the deep semantic space faster than IR.

\begin{figure*}[h]
   \centering
   \includegraphics[width=1\linewidth]{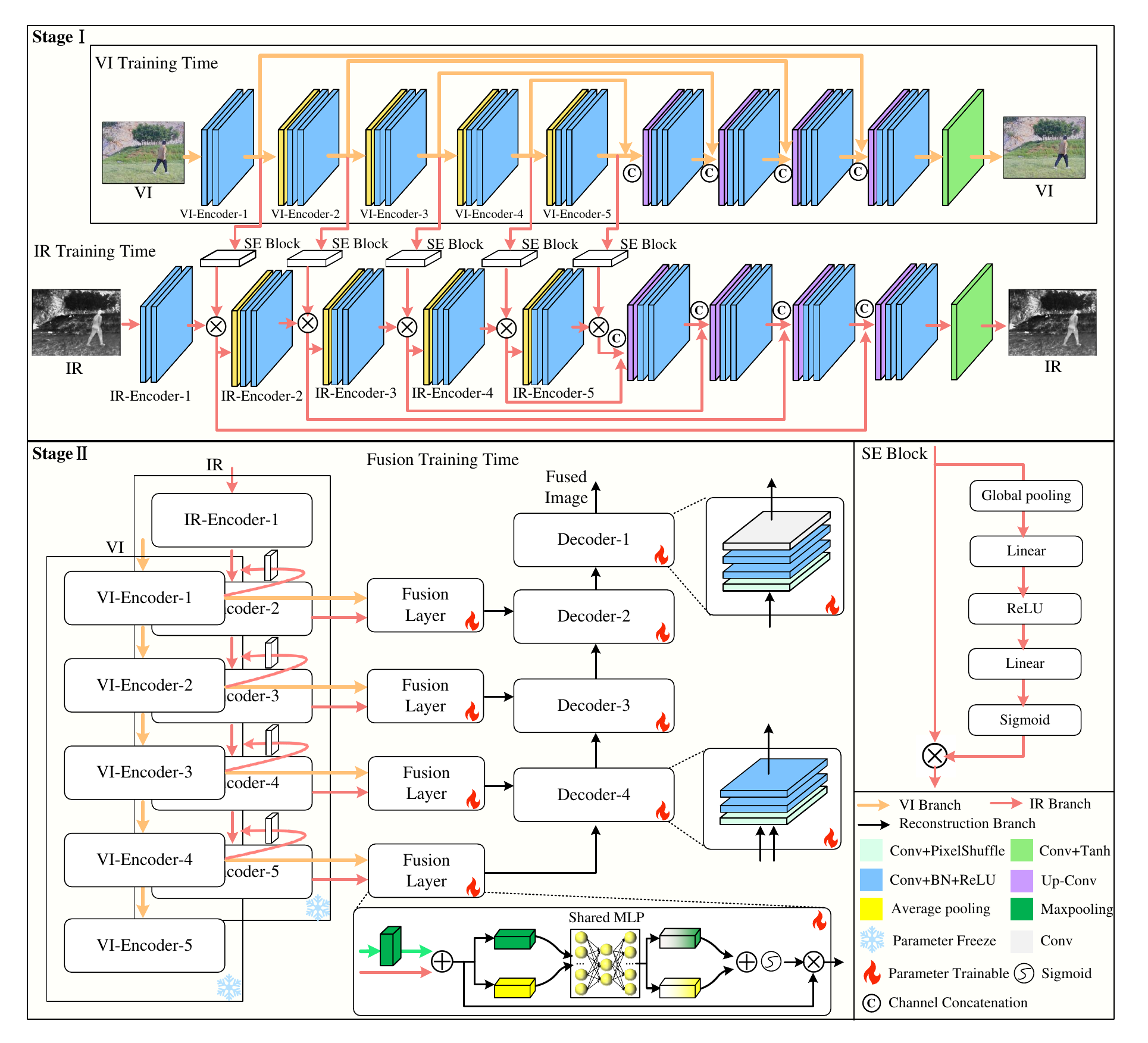}
   \caption{Workflow of MMA-UNet.}
   \label{fig2}
\end{figure*}

\section{Related Work}
\textbf{Infrared and Visible Image Fusion Methods.}
The existing deep learning-based IVIF algorithms are primarily divided into two categories: generative and encoding models. Generative models are primarily based on the generative adversarial network (GAN) \cite{r27,r28,r29,r54,r58} and diffusion \cite{r30,r31}. They aimed to learn data distribution from the latent space and simulate the distribution of the target data by generating the data. For example, Liu et al. \cite{r19} proposed a target-aware dual adversarial learning strategy to generate fusion results that are beneficial for downstream tasks. They designed the target and detailed discriminators to learn the target features from IR and VI, respectively. To address the issues of weak training stability and lack of interpretability in GAN-based models, Zhao et al. \cite{r30} introduced a denoising diffusion model for IVIF, defining the fusion task as an unconditional generation sub-problem and a maximum likelihood sub-problem. Conversely, encoding models do not generate new data samples and instead, extract key features from the original input data, mapping the original data to a more compact representation space \cite{r32,r33,r34,r35,r36,r37,r38}. For instance, Ma et al. \cite{r32} introduced Swin Transformer to the image fusion, incorporating an attention-guided cross-domain module. This design effectively integrates multi-modality complementary information and global interaction information. To enhance the interpretability of feature extraction, Li et al. \cite{r17} proposed an image fusion framework based on representation learning, that established a connection between the mathematical formulas and network architecture to enhance the interpretability of feature extraction. They utilized low-rank representation learning theory to establish an image decomposition model, effectively avoiding a time-consuming network design. Owing to the continuous improvement in image fusion frameworks, several researchers have begun to analyze the performance of fusion models in promoting downstream tasks. The many multitask joint learning methods have emerged \cite{r19,r33}. For example, fusion and object detection \cite{r19}, fusion and semantic segmentation \cite{r33,r39}, fusion and salient object detection \cite{r40}, and fusion and low-light enhancement \cite{r41}. They generally guide the learning of fusion networks through the performance feedback of the fusion results on downstream tasks. Additionally, several works proposed "registration and fusion" models \cite{r18,r42,r43} due to the spatial deformation and misalignment in the imaging process of multi-modal sensors to effectively avoid the excessive reliance of algorithms on the registration data pairs.

\section{Method and Analysis}
In this section, we first provide details of all components of MMA-UNet and loss functions. Then, the principle and feasibility of asymmetric architecture design and guidance mechanism are further analyzed.

\subsection{IR-UNet and VI-UNet}
In the stage \Rmnum{1}, two different UNet models were trained separately using IR and VI. Note that IR-UNet was trained first, then VI-UNet train.As shown in  \cref{fig2}, to train VI-UNet, we introduced the original UNet architecture \cite{r44} and fine-tuned it by converting the input and output into three channels. To train IR-UNet, we utilized the information from VI to assist IR-UNet in learning the IR features, thereby accelerating the feature extraction process of the network. Specifically, a part of autoencoder is consistent with the VI-UNet. The only difference lies in the feature extraction stage of the encoder, where we employ the Squeeze-and-Excitation (SE Block) module \cite{r45} to obtain the attention maps of the VI information, which are then element-wise multiplied by the IR feature maps. This process can inject significant information captured by VI into the IR features to assist the IR-UNet network in extracting the features. Notably, the parameters of VI-UNet were frozen during the training of IR-UNet. To train these two UNet models separately, the mean squared error (MSE) was introduced as the loss function to achieve image decomposition and reconstruction using the following formula:
\begin{equation}
{Loss}_{mse}\left(X,\ O\right)=\frac{1}{HW}\sum_{i=1}^{H}\sum_{j=1}^{W}{{\left[X\left(i,j\right)-O\left(i,j\right)\right]\ }^2}
\label{eq1}
\end{equation}
where $H$ and $W$ represent the height and width of the image, respectively, $X$ represents the input image, and $O$ represents the output image.

\subsection{Asymmetric UNet}
The stage \Rmnum{2} involves fusion and reconstruction. Contrary to the vanilla fusion manner \cite{r46}, we adopted an asymmetric architecture. Since different modal images have unique information space distributions, the number of the convolution layers for extracting the deep semantic features should also vary for different modalities within the same architecture. Therefore, we designed an asymmetric UNet architecture by combining the CKA similarity results to enable the fusion of features from different modalities with same information space distributions. Specifically, as shown in  \cref{fig2}, we employed the encoders of the IR-UNet and VI-UNet to extract the corresponding low-level and deep semantic features of IR and VI. Subsequently, we fused the features from the first four layers of the VI-UNet with the last four layers of the IR-UNet model. For example, we fused the first four layers of the features of VI-UNet with the last four layers of the features of IR-UNet to generate four sets of fused feature maps. For the fusion strategy, we first down-sampled the feature maps of VI to match the size of the feature maps of IR. We then performed feature addition and implemented a channel-attention operation \cite{r47} for the added features. This amplifies the important features and suppresses the irrelevant ones. Lastly, we reconstructed the fused features to obtain a fused image. For the first three layers of the decoder, we used the convolution and pixelshuffle operations for up-sampling. The last layer of the decoder added a convolution layer with a $1 \times 1$ kernel to the original basis.
In the second stage, we introduced the MSE, structural similarity index measure(SSIM), and L1 norms to calculate the relevant loss functions. The SSIM is calculated as follows:
\begin{equation}
SSIM\left(X,\ Y\right)=\frac{\left(2\mu_X\mu_Y+C_1\right)\left(2\sigma_{XY}+C_2\right)}{\left(\mu_X^2+\mu_Y^2+C_1\right)\left(\sigma_X^2+\sigma_Y^2+C_2\right)}
\label{eq2}
\end{equation}
where $Y$ and $Z$ represent two different images. $\mu_X$ represents the mean of image $X$, $\sigma_X$ represents the standard deviation of image $X$, and $\sigma_{XY}$ represents the covariance of images, $X$ and $Y$. $C1$ and $C2$ represent the constants used to prevent the denominator in the formula from being close to $0$.
Therefore, the calculation formula of the structural loss, $loss_{ssim}$, is as follows:
\begin{equation}
{Loss}_{ssim}\left(X,Y,Z\right)=SSIM\left(Z,X\right)+SSIM\left(Z,\ Y\right)
\label{eq3}
\end{equation}
The detail loss, $loss_{det}$, is expressed as follows:
\begin{equation}
{Loss}_{det}\left(X, Y, Z\right) = \nabla Z - \max(\nabla X, \nabla Y)^2 \quad
\label{eq4}
\end{equation}
The \(\nabla\) is a sobel operator.

The total loss function, $Loss$, can be expressed as follows:
\begin{align}
\textit{Loss} & = \textit{Loss}_{\textit{mse}}\left(\left(\text{IR}+\text{VI}\right)/2, \text{F}\right) \notag \\
& + \alpha \textit{Loss}_{\textit{ssim}}\left(\text{IR}, \text{VI}, \text{F}\right) + \beta \textit{Loss}_{\textit{det}}\left(\text{IR}, \text{VI}, \text{F}\right)
\label{eq5}
\end{align}
where F represents the fused image.

\subsection{Analysis for Multi-modal Feature Extraction}
\begin{figure*}[h]
   \centering
   \includegraphics[width=0.75\linewidth]{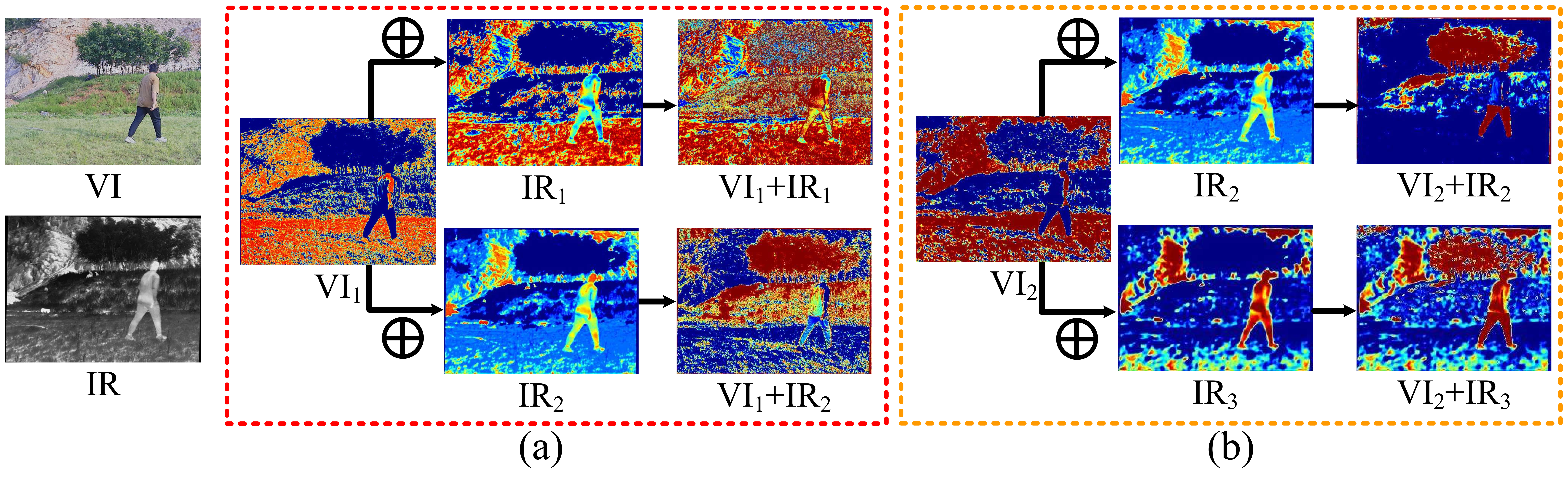}
   \caption{VI1 represents the intermediate feature representation of the first convolutional layer in UNet. The same applies to IR1, IR2, IR3 and VI2. VI1+IR1 indicates that two features are added together to obtain a fusion map. VI1+IR2, VI2+IR2, VI2+IR3, likewise. For ease of representation, we omit the sampling operation.}
   \label{fig3}
\end{figure*}

Previous studies \cite{r38,r52,r53} did not discuss the differences in the rates of feature extraction of the different modalities under the same framework. To the end, we performed a visual analysis of the feature maps of VI-UNet and IR-UNet.  \cref{fig3}(a) presents the addition results of the feature maps of the different layers. In the shallow layer, the encoder extracts pixel-level features, such as the texture of leaves and grass and details of pedestrians. Simultaneously, this shallow information can be recognized in different modal images. When the same layers of two modalities perform an addition operation, the result becomes redundant. For MMIF tasks, it is crucial to focus on the complementary information obtained from different modalities. For common features, priority should be given to the extraction of pixels captured by the VI sensors, since they correspond more closely to the human visual system (HVS). Additionally, handling large numbers of similar features from multiple modalities weakens the representational capacity of the model, thereby affecting its performance and generalization. Conversely, asymmetric fusion corresponds to the concept of MMIF that integrate useful information from different modal image and eliminate redundant information. It integrates the IR radiation information while preserving the VI details. The features extracted by the encoder became more abstract as the number of layers increased. There is an imbalance in the feature extraction depth, where features extracted from one modality are more abstract, while those from another modality are relatively shallow. Therefore, direct fusion may lead to the model relying excessively on the abstract features, resulting in information imbalance. The fusion results at the same layer lose feature information from the IR, whereas in the results obtained from the asymmetric fusion strategy, the complementary features from multiple modalities are well preserved, as shown in  \cref{fig3}(b).

\subsection{Analysis for Guidance Mechanism}
\begin{figure}[h]
   \centering
   \includegraphics[width=1\linewidth]{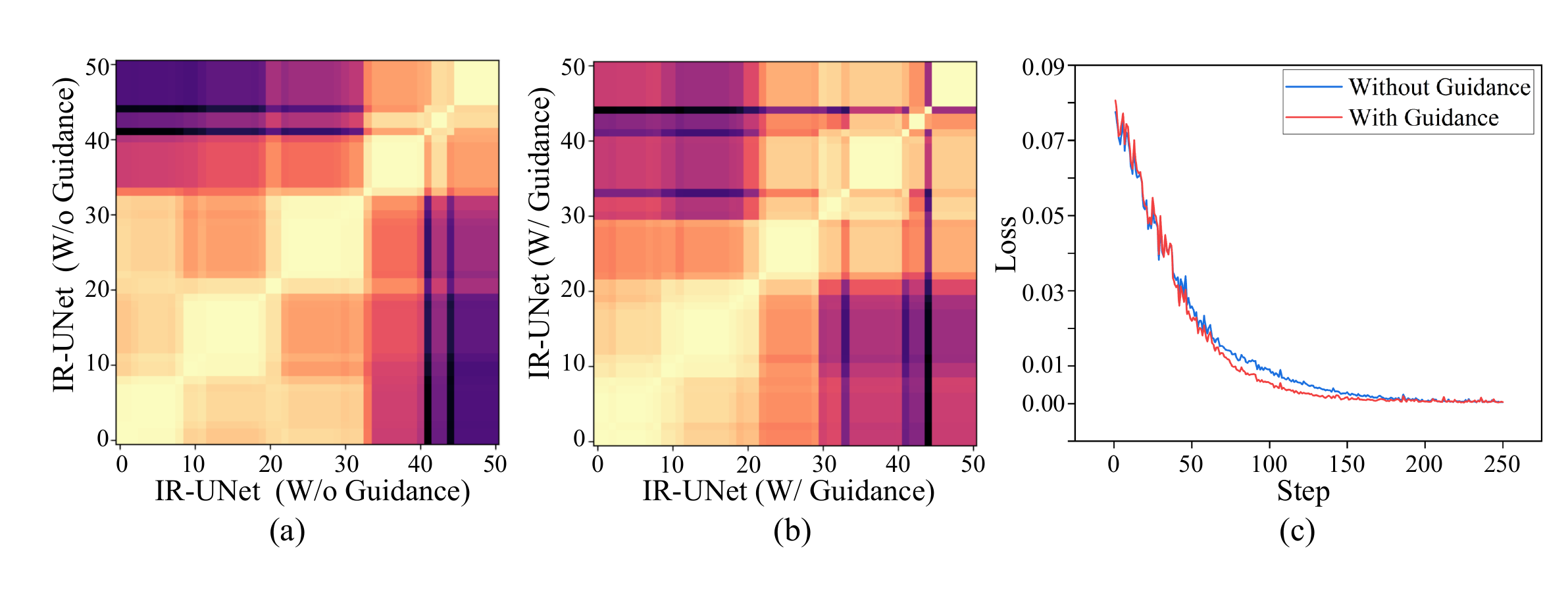}
   \caption{Centered Kernel Alignment. (a) and (b) represent the computation of the CKA similarity between all pairs of layers in IR-UNet with and without guidance mechanism, respectively.}
   \label{fig4}
\end{figure}

There are differences in the speed of extracting deep semantic features within the same framework owing to the inconsistent spatial information distribution from different modalities. To expedite the extraction of deep semantic information from the IR, we utilized VI features to guide the reconstruction of the IR features at each layer. We compared the two sets of CKA similarities obtained using two different training methods. IR-UNet without the guidance mechanism exhibited significant semantic information divergence only after the 30th layer, as shown in  \cref{fig4}. However, by introducing the guidance, IR-UNet began to exhibit deep semantic features after the 22nd layer. This experimental phenomenon demonstrates that the guidance mechanism facilitates cross-modal knowledge transfer, enabling IR-UNet to learn task-relevant semantic features more rapidly and accelerates the overfitting speed of the model.

\begin{figure}[h]
   \centering
   \includegraphics[width=1\linewidth]{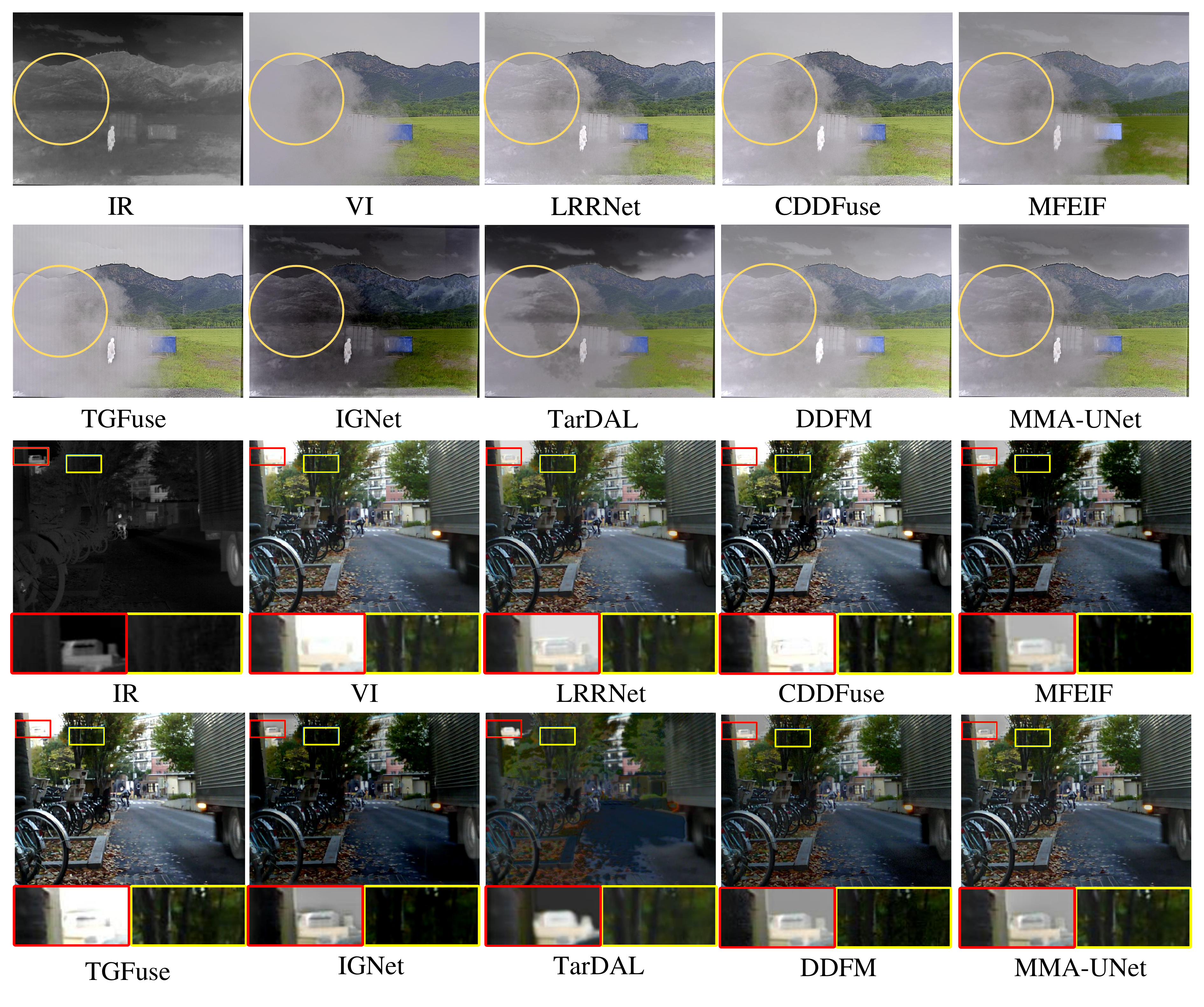}
   \caption{Subjective comparisons of fusion results obtained by MMA-UNet and the SoTA comparison methods on M3FD and MSRS.}
   \label{fig5}
\end{figure}

\begin{figure*}[t]
   \centering
   \includegraphics[width=1\linewidth]{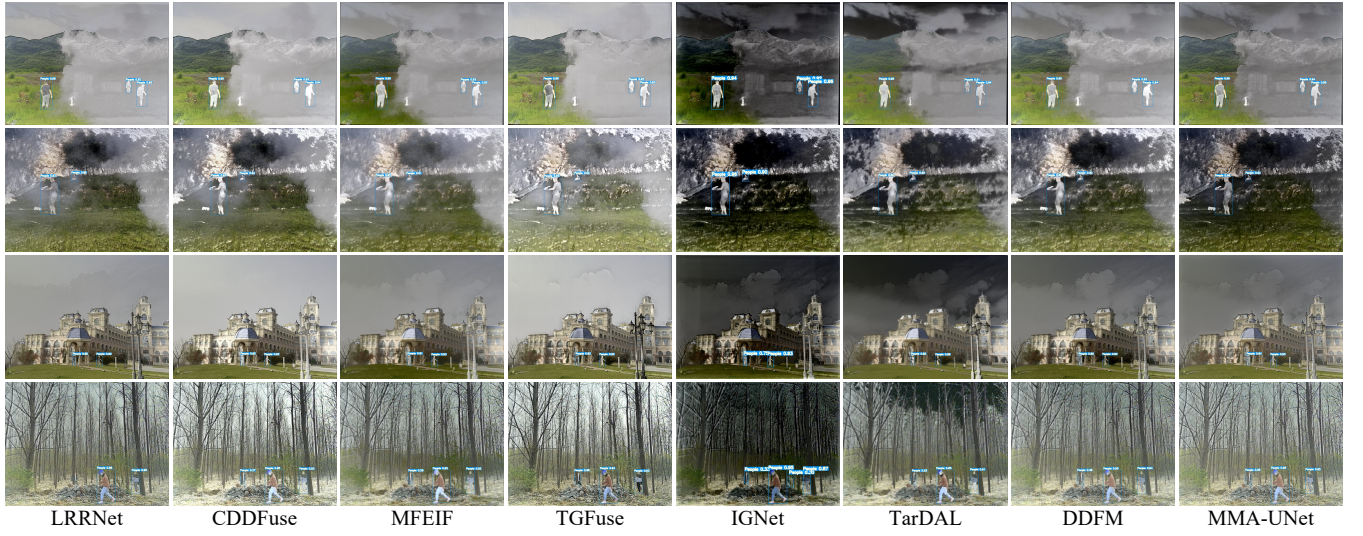}
   \caption{Results of all methods on the detection task.}
   \label{fig6}
\end{figure*}

\begin{figure*}[t]
   \centering
   \includegraphics[width=1\linewidth]{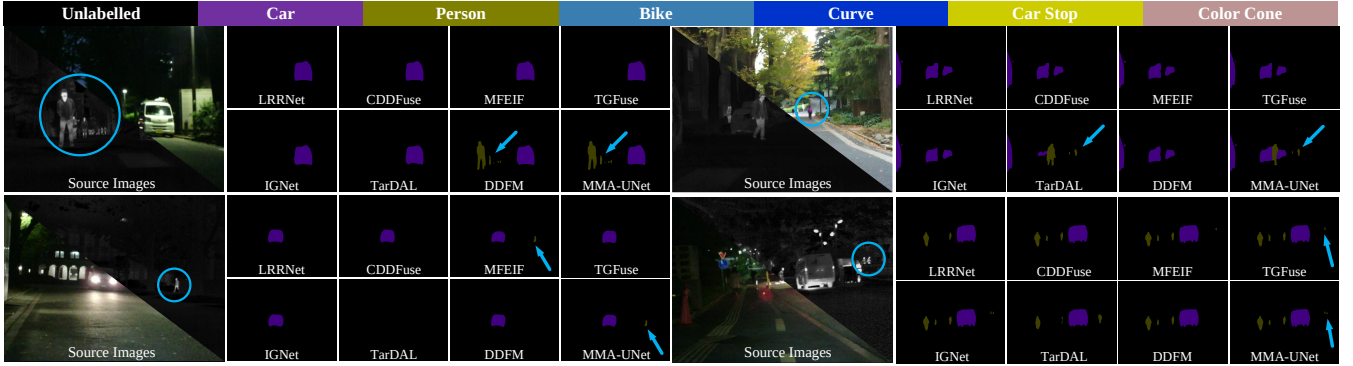}
   \caption{The segmentation accuracy of the comparison methods and MMA-UNet on various categories.}
   \label{fig7}
\end{figure*}

\section{Experiment}
\subsection{Experimental Setting}
Experiments were performed on two NVIDIA GeForce RTX 3090 GPUs with a 64-core Intel Xeon Platinum 8350C CPU. The experimental settings of VI-UNet and IR-UNet are the same. We selected AdamW optimizer to adjust the training parameters. The basic learning rate was initially set to $1e-3$, and the weight decay was set to $0.005$. The cosine annealing strategy was adopted to adjust learning rate adaptively. In terms of data enhancement, we adopted the random resized crop. In addition, the experimental settings of the fusion framework are only slightly different from IR-UNet. The basic learning rate was changed to $1e-4$. In the loss function, $\alpha$ and $\beta$ are set as $10$, $0.5$ respectively.

\subsection{Dataset and Evaluation Metrics}
We choose M3FD \cite{r19} and MSRS \cite{r48} as the datasets for the experiment. Specifically, we selected $4200$ and $1083$ pairs of images from M3FD and MSRS respectively as the training set. As for the testing set, there are $300$ M3FD pairs and $361$ MSRS pairs.

In order to verify the superiority of the fusion performance of MMA-UNet, we selected seven state-of-the-art (SoTA) comparison methods. These comparison methods encompass all popular frameworks for MMIF, including algorithmic unfolding models (LRRNet) \cite{r17}, Hybrid model (CDDFuse) \cite{r3}, CNN-based model (MFEIF) \cite{r36}, Transformer-Based model (TGFuse) \cite{r34}, GNN-based model (IGNet) \cite{r5}, GAN-Based model (TarDAL) \cite{r19}, Diffusion-Based and model (DDFM) \cite{r30}. 

The five popular objective metrics were selected as qualitative evaluation of different fusion methods, including Chen-Blum Metric ($Q_{cb}$) , edge-based similarity measure ($Q_{abf}$), visual information fidelity (VIF), SSIM and peak signal-to-noise ratio(PSNR) \cite{r1,r49}. For these metrics, higher scores represent higher quality of the fused image.

In the detection task, we used the labeled image pairs provided by M3FD and divide them into training, validation and test set according to 6:3:1. The mainstream detection network, YOLOv7 \cite{r7}, was employed to detect fusion results. In the segmentation task, we utilized the training and test set provided by the MSRS to conduct the training and test experiments of the segmentation network \cite{r50}. To ensure experimental rigor, the experimental settings of all downstream task models strictly follow the conditions provided in the original text.

\begin{table*}[h]
\caption{Quantitative comparison of MMA-UNet with seven SoTA methods on the M3FD and MSRS datasets. The first, second, and third rankings are represented by the red, blue and green fonts, respectively.}
\centering
\begin{tabular}{c|c|ccccc|ccccc}
\hline
 &    & \multicolumn{5}{c|}{Dataset: M3FD} & \multicolumn{5}{c}{Dataset: MSRS}   \\ \cline{3-12} 
\multirow{-2}{*}{Methods}                & \multirow{-2}{*}{Pub.} & \multicolumn{1}{c|}{VIF ↑}                          & \multicolumn{1}{c|}{SSIM ↑}                         & \multicolumn{1}{c|}{PSNR ↑}                           & \multicolumn{1}{c|}{$Q_{abf}$ ↑}                         & $Q_{cb}$ ↑                         & \multicolumn{1}{c|}{SSIM ↑}                         & \multicolumn{1}{c|}{VIF ↑}                          & \multicolumn{1}{c|}{PSNR ↑}                          & \multicolumn{1}{c|}{$Q_{abf}$ ↑}                          & $Q_{cb}$ ↑                         \\ \hline
\multicolumn{1}{c|}{ Algorithmic unfolding \cite{r17}} & TPAMI 23               & \multicolumn{1}{c|}{0.378}                        & \multicolumn{1}{c|}{0.749}                        & \multicolumn{1}{c|}{15.458}                        & \multicolumn{1}{c|}{0.563}                        & 0.487                        & \multicolumn{1}{c|}{0.364}                        & \multicolumn{1}{c|}{{\color[HTML]{00B0F0} 0.564}} & \multicolumn{1}{c|}{11.501}                        & \multicolumn{1}{c|}{0.638}                        & 0.472                        \\
\multicolumn{1}{c|}{Hybrid-based \cite{r3}}        & CVPR 23                & \multicolumn{1}{c|}{0.392}                        & \multicolumn{1}{c|}{0.725}                        & \multicolumn{1}{c|}{13.446}                        & \multicolumn{1}{c|}{{\color[HTML]{00B0F0} 0.672}} & {\color[HTML]{32CB00} 0.511} & \multicolumn{1}{c|}{0.288}                        & \multicolumn{1}{c|}{0.379}                        & \multicolumn{1}{c|}{12.802}                        & \multicolumn{1}{c|}{0.433}                        & 0.398                        \\
\multicolumn{1}{c|}{CNN-based \cite{r36}}           & TCSVT 21               & \multicolumn{1}{c|}{{\color[HTML]{32CB00} 0.424}} & \multicolumn{1}{c|}{{\color[HTML]{00B0F0} 0.763}} & \multicolumn{1}{c|}{{\color[HTML]{32CB00} 15.630}} & \multicolumn{1}{c|}{0.582}                        & 0.469                        & \multicolumn{1}{c|}{{\color[HTML]{00B0F0} 0.420}} & \multicolumn{1}{c|}{0.547}                        & \multicolumn{1}{c|}{10.953}                        & \multicolumn{1}{c|}{{\color[HTML]{FF0000} 0.756}} & 0.464                        \\
\multicolumn{1}{c|}{Transformer-based \cite{r34}}   & TIP 23                 & \multicolumn{1}{c|}{0.382}                        & \multicolumn{1}{c|}{0.720}                        & \multicolumn{1}{c|}{13.241}                        & \multicolumn{1}{c|}{{\color[HTML]{FE0000} 0.681}} & {\color[HTML]{00B0F0} 0.527} & \multicolumn{1}{c|}{0.280}                        & \multicolumn{1}{c|}{0.414}                        & \multicolumn{1}{c|}{{\color[HTML]{32CB00} 12.927}} & \multicolumn{1}{c|}{0.500}                        & 0.405                        \\
\multicolumn{1}{c|}{GNN-based \cite{r50}}           & MM 23                  & \multicolumn{1}{c|}{0.254}                        & \multicolumn{1}{c|}{0.609}                        & \multicolumn{1}{c|}{13.088}                        & \multicolumn{1}{c|}{0.446}                        & 0.424                        & \multicolumn{1}{c|}{{\color[HTML]{32CB00} 0.396}} & \multicolumn{1}{c|}{{\color[HTML]{32CB00} 0.554}} & \multicolumn{1}{c|}{10.800}                        & \multicolumn{1}{c|}{{\color[HTML]{00B0F0} 0.731}} & {\color[HTML]{00B0F0} 0.482} \\
\multicolumn{1}{c|}{GAN-based \cite{r19}}           & CVPR 22                & \multicolumn{1}{c|}{{\color[HTML]{FF0000} 0.462}} & \multicolumn{1}{c|}{{\color[HTML]{32CB00} 0.760}} & \multicolumn{1}{c|}{14.840}                        & \multicolumn{1}{c|}{0.246}                        & 0.434                        & \multicolumn{1}{c|}{0.254}                        & \multicolumn{1}{c|}{0.467}                        & \multicolumn{1}{c|}{12.317}                        & \multicolumn{1}{c|}{0.125}                        & 0.379                        \\
\multicolumn{1}{c|}{Diffusion-based \cite{r30}}     & ICCV 23                & \multicolumn{1}{c|}{0.423}                        & \multicolumn{1}{c|}{0.756}                        & \multicolumn{1}{c|}{{\color[HTML]{00B0F0} 16.334}} & \multicolumn{1}{c|}{0.529}                        & 0.478                        & \multicolumn{1}{c|}{0.352}                        & \multicolumn{1}{c|}{0.510}                        & \multicolumn{1}{c|}{{\color[HTML]{FF0000} 13.485}} & \multicolumn{1}{c|}{0.543}                        & {\color[HTML]{32CB00} 0.479} \\ \hline
\multicolumn{1}{c|}{ MMA-UNet}            &                        & \multicolumn{1}{c|}{{\color[HTML]{00B0F0} 0.430}} & \multicolumn{1}{c|}{{\color[HTML]{FF0000} 0.764}} & \multicolumn{1}{c|}{{\color[HTML]{FF0000} 16.335}} & \multicolumn{1}{c|}{{\color[HTML]{32CB00} 0.626}} & {\color[HTML]{FF0000} 0.529} & \multicolumn{1}{c|}{{\color[HTML]{FF0000} 0.440}} & \multicolumn{1}{c|}{{\color[HTML]{FF0000} 0.576}} & \multicolumn{1}{c|}{{\color[HTML]{00B0F0} 13.214}} & \multicolumn{1}{c|}{{\color[HTML]{32CB00} 0.699}} & {\color[HTML]{FF0000} 0.502} \\ \hline
\end{tabular}
\label{tab1}
\end{table*}

\begin{table*}[h]
\caption{The detection accuracy of the comparison methods and MMA-UNet on various categories in the M3FD dataset. The first, second, and third rankings are represented by the red, blue and green fonts, respectively.}
\centering
\begin{adjustbox}{width=\textwidth}
\begin{tabular}{c|ccccccc|ccccccc}
\hline
 & \multicolumn{7}{c|}{AP@0.5}    & \multicolumn{7}{c}{AP@{[}0.5:0.95{]}}   \\ \cline{2-15} 
\multirow{-2}{*}{Methods} & \multicolumn{1}{c|}{People}                       & \multicolumn{1}{c|}{Car}                          & \multicolumn{1}{c|}{Bus}                          & \multicolumn{1}{c|}{Motorcycle}                   & \multicolumn{1}{c|}{Lamp}                         & \multicolumn{1}{c|}{Truck}                        & mAP                          & \multicolumn{1}{c|}{People}                       & \multicolumn{1}{c|}{Car}                          & \multicolumn{1}{c|}{Bus}                          & \multicolumn{1}{c|}{Motorcycle}                   & \multicolumn{1}{c|}{Lamp}                         & \multicolumn{1}{c|}{Truck}                        & mAP                          \\ \hline
Infrared image                        & \multicolumn{1}{c|}{0.787}                        & \multicolumn{1}{c|}{0.760}                        & \multicolumn{1}{c|}{0.669}                        & \multicolumn{1}{c|}{0.023}                        & \multicolumn{1}{c|}{0.400}                        & \multicolumn{1}{c|}{0.386}                        & 0.504                        & \multicolumn{1}{c|}{0.445}                        & \multicolumn{1}{c|}{0.477}                        & \multicolumn{1}{c|}{0.433}                        & \multicolumn{1}{c|}{0.004}                        & \multicolumn{1}{c|}{0.214}                        & \multicolumn{1}{c|}{0.205}                        & 0.296                        \\
Visible image                      & \multicolumn{1}{c|}{0.374}                        & \multicolumn{1}{c|}{0.868}                        & \multicolumn{1}{c|}{0.760}                        & \multicolumn{1}{c|}{0.746}                        & \multicolumn{1}{c|}{0.536}                        & \multicolumn{1}{c|}{{\color[HTML]{32CB00} 0.902}} & 0.698                        & \multicolumn{1}{c|}{0.158}                        & \multicolumn{1}{c|}{0.572}                        & \multicolumn{1}{c|}{0.519}                        & \multicolumn{1}{c|}{0.388}                        & \multicolumn{1}{c|}{0.267}                        & \multicolumn{1}{c|}{{\color[HTML]{32CB00} 0.666}} & 0.428                        \\
 Algorithmic unfolding \cite{r17}        & \multicolumn{1}{c|}{0.824}                        & \multicolumn{1}{c|}{0.928}                        & \multicolumn{1}{c|}{0.897}                        & \multicolumn{1}{c|}{{\color[HTML]{FF0000} 0.790}} & \multicolumn{1}{c|}{{\color[HTML]{FF0000} 0.662}} & \multicolumn{1}{c|}{0.873}                        & {\color[HTML]{00B0F0} 0.829} & \multicolumn{1}{c|}{0.458}                        & \multicolumn{1}{c|}{{\color[HTML]{32CB00} 0.672}} & \multicolumn{1}{c|}{0.635}                        & \multicolumn{1}{c|}{{\color[HTML]{00B0F0} 0.423}} & \multicolumn{1}{c|}{0.344}                        & \multicolumn{1}{c|}{0.656}                        & 0.531                        \\
Hybrid-based \cite{r3}              & \multicolumn{1}{c|}{0.830}                        & \multicolumn{1}{c|}{{\color[HTML]{32CB00} 0.932}} & \multicolumn{1}{c|}{0.855}                        & \multicolumn{1}{c|}{0.780}                        & \multicolumn{1}{c|}{0.581}                        & \multicolumn{1}{c|}{{\color[HTML]{FF0000} 0.919}} & 0.816                        & \multicolumn{1}{c|}{0.477}                        & \multicolumn{1}{c|}{0.664}                        & \multicolumn{1}{c|}{{\color[HTML]{32CB00} 0.657}} & \multicolumn{1}{c|}{0.396}                        & \multicolumn{1}{c|}{0.330}                        & \multicolumn{1}{c|}{{\color[HTML]{00B0F0} 0.667}} & 0.532                        \\
CNN-based \cite{r36}                 & \multicolumn{1}{c|}{{\color[HTML]{32CB00} 0.848}} & \multicolumn{1}{c|}{{\color[HTML]{FF0000} 0.940}} & \multicolumn{1}{c|}{0.845}                        & \multicolumn{1}{c|}{0.782}                        & \multicolumn{1}{c|}{0.600}                        & \multicolumn{1}{c|}{0.894}                        & 0.818                        & \multicolumn{1}{c|}{{\color[HTML]{32CB00} 0.488}} & \multicolumn{1}{c|}{0.671}                        & \multicolumn{1}{c|}{0.639}                        & \multicolumn{1}{c|}{0.410}                        & \multicolumn{1}{c|}{{\color[HTML]{32CB00} 0.349}} & \multicolumn{1}{c|}{0.608}                        & 0.528                        \\
Transformer-based \cite{r34}         & \multicolumn{1}{c|}{0.832}                        & \multicolumn{1}{c|}{0.934}                        & \multicolumn{1}{c|}{0.897}                        & \multicolumn{1}{c|}{{\color[HTML]{32CB00} 0.783}} & \multicolumn{1}{c|}{0.573}                        & \multicolumn{1}{c|}{0.888}                        & 0.818                        & \multicolumn{1}{c|}{0.474}                        & \multicolumn{1}{c|}{0.666}                        & \multicolumn{1}{c|}{0.637}                        & \multicolumn{1}{c|}{0.412}                        & \multicolumn{1}{c|}{0.340}                        & \multicolumn{1}{c|}{{\color[HTML]{FF0000} 0.667}} & {\color[HTML]{32CB00} 0.533} \\
GNN-based \cite{r50}                 & \multicolumn{1}{c|}{0.835}                        & \multicolumn{1}{c|}{0.922}                        & \multicolumn{1}{c|}{{\color[HTML]{FF0000} 0.950}} & \multicolumn{1}{c|}{0.659}                        & \multicolumn{1}{c|}{{\color[HTML]{00B0F0} 0.661}} & \multicolumn{1}{c|}{0.873}                        & 0.817                        & \multicolumn{1}{c|}{0.479}                        & \multicolumn{1}{c|}{0.656}                        & \multicolumn{1}{c|}{{\color[HTML]{00B0F0} 0.674}} & \multicolumn{1}{c|}{0.291}                        & \multicolumn{1}{c|}{0.330}                        & \multicolumn{1}{c|}{0.614}                        & 0.507                        \\
GAN-based \cite{r19}                 & \multicolumn{1}{c|}{0.757}                        & \multicolumn{1}{c|}{0.823}                        & \multicolumn{1}{c|}{0.807}                        & \multicolumn{1}{c|}{0.586}                        & \multicolumn{1}{c|}{0.500}                        & \multicolumn{1}{c|}{{\color[HTML]{00B0F0} 0.917}} & 0.732                        & \multicolumn{1}{c|}{0.449}                        & \multicolumn{1}{c|}{0.625}                        & \multicolumn{1}{c|}{{\color[HTML]{FF0000} 0.676}} & \multicolumn{1}{c|}{0.328}                        & \multicolumn{1}{c|}{0.339}                        & \multicolumn{1}{c|}{0.594}                        & 0.502                        \\
Diffusion-based \cite{r30}           & \multicolumn{1}{c|}{{\color[HTML]{FF0000} 0.851}} & \multicolumn{1}{c|}{0.930}                        & \multicolumn{1}{c|}{{\color[HTML]{32CB00} 0.901}} & \multicolumn{1}{c|}{0.753}                        & \multicolumn{1}{c|}{{\color[HTML]{32CB00} 0.643}} & \multicolumn{1}{c|}{0.856}                        & {\color[HTML]{32CB00} 0.822} & \multicolumn{1}{c|}{{\color[HTML]{00B0F0} 0.493}} & \multicolumn{1}{c|}{{\color[HTML]{00B0F0} 0.676}} & \multicolumn{1}{c|}{0.644}                        & \multicolumn{1}{c|}{{\color[HTML]{32CB00} 0.422}} & \multicolumn{1}{c|}{{\color[HTML]{FF0000} 0.377}} & \multicolumn{1}{c|}{0.617}                        & {\color[HTML]{00B0F0} 0.538} \\ \hline
MMA-UNet                  & \multicolumn{1}{c|}{{\color[HTML]{00B0F0} 0.850}} & \multicolumn{1}{c|}{{\color[HTML]{00B0F0} 0.939}} & \multicolumn{1}{c|}{{\color[HTML]{00B0F0} 0.911}} & \multicolumn{1}{c|}{{\color[HTML]{00B0F0} 0.786}} & \multicolumn{1}{c|}{0.613}                        & \multicolumn{1}{c|}{0.883}                        & {\color[HTML]{FF0000} 0.830} & \multicolumn{1}{c|}{{\color[HTML]{FF0000} 0.496}} & \multicolumn{1}{c|}{{\color[HTML]{FF0000} 0.685}} & \multicolumn{1}{c|}{0.639}                        & \multicolumn{1}{c|}{{\color[HTML]{FF0000} 0.425}} & \multicolumn{1}{c|}{{\color[HTML]{00B0F0} 0.356}} & \multicolumn{1}{c|}{0.658}                        & {\color[HTML]{FF0000} 0.543} \\ \hline
\end{tabular}
\end{adjustbox}
\label{tab2}
\end{table*}
\begin{table*}[h]
\caption{The segmentation accuracy of the comparison methods and MMA-UNet on various categories in the MSRS dataset. The first, second, and third rankings are represented by the red, blue and green fonts, respectively.}
\centering
\begin{tabular}{c|c|c|c|c|c|c|c|c}
\hline
Method              & Background                   & Car                          & Person                       & Bike                         & Curve                        & Car Stop                     & Color Cone                   & mIoU                         \\ \hline
Infrared image                & 94.45                        & 58.67                        & 80.63                        & 18.46                        & 8.67                         & 8.21                         & 0.00                         & 38.44                        \\
Visible image                 & 97.40                        & 87.36                        & 40.78                        & 82.34                        & {\color[HTML]{32CB00} 66.04} & {\color[HTML]{FF0000} 55.55} & {\color[HTML]{00B0F0} 48.13} & 68.23                        \\
 Algorithmic unfolding \cite{r17} & 97.97                        & 87.10                        & 73.96                        & 82.38                        & 64.57                        & 51.57                        & {\color[HTML]{32CB00} 46.63} & 72.03                        \\
Hybrid-based \cite{r3}        & {\color[HTML]{FF0000} 98.23} & {\color[HTML]{FF0000} 88.86} & 81.16                        & {\color[HTML]{32CB00} 82.70} & {\color[HTML]{00B0F0} 66.06} & {\color[HTML]{00B0F0} 55.03} & 46.45                        & {\color[HTML]{32CB00} 74.07} \\
CNN-based \cite{r36}           & 98.12                        & 88.03                        & {\color[HTML]{00B0F0} 82.33} & 81.57                        & {\color[HTML]{FF0000} 66.78}                        & 42.99                        & 26.19                        & 69.43                        \\
Transformer-based \cite{r34}   & {\color[HTML]{00B0F0} 98.23}                        & {\color[HTML]{00B0F0} 88.69}                        & {\color[HTML]{FF0000} 82.37}                        & {\color[HTML]{FF0000} 82.9}                          & 65.98                        & 54.09                        & 46.27                        & {\color[HTML]{00B0F0} 74.08}                        \\
GNN-based \cite{r50}           & 98.00                        & 87.13                        & {\color[HTML]{32CB00} 82.04}                        & 79.77                        & 64.5                         & 35.74                        & 44.09                        & 70.18                        \\
GAN-based \cite{r19}           & 97.18                        & 81.63                        & 63.94                        & 75.43                        & 38.81                        & 30.22                        & 0.00                         & 55.32                        \\
Diffusion-based \cite{r30}     & 98.18                        & 88.44                        & 81.98                        & 82.08                        & 65.64                        & 51.49                        & 46.48                        & 73.47                        \\ \hline
MMA-UNet            & {\color[HTML]{32CB00} 98.20}                        & {\color[HTML]{32CB00} 88.61}                        & 80.55                        & {\color[HTML]{00B0F0} 82.77}                        & 65.92                        & {\color[HTML]{32CB00} 54.39}                        & {\color[HTML]{FF0000} 48.42}                        & {\color[HTML]{FF0000} 74.12}                        \\ \hline
\end{tabular}
\label{tab3}
\end{table*}

\subsection{Qualitative Analysis}
\textit{Fusion analysis.} As the  \cref{fig5} shows, MMA-UNet is better than the comparison methods in terms of detail and structural information. For example, in the region marked by a circle in the first set of fused images, when obscured by thick smoke, MMA-UNet can well preserve the edge contours and detailed texture information of the mountains captured by the IR. In addition, in the second set of fusion result, when the VI is overexposed, MMA-UNet can well preserve the structure of the house in the IR and eliminate the overexposure. This is all owe to asymmetric fusion of the same feature space and achieving information balance in different modalities. Lastly, IGNet, MFEIF and DDFM all have varying degrees of loss of detail information and reduced contrast in the two fusion results.

\textit{Detection analysis.}  \cref{fig6} illustrates the comparison of the MMA-UNet and the comparison methods in detection experiments. In this scenario, smoke in the VI obscures the pedestrian information inside. Therefore, the fusion method needs to identify pixels that extract interference features and capture the most valuable thermal energy information from the IR. As shown in the  \cref{fig6}, the detector achieved the highest accuracy with the MMA-UNet, indicating the promotional effect of the proposed algorithm on downstream tasks.

\textit{Segmentation analysis.}  \cref{fig7} displays the semantic segmentation results of the fused images generated by different methods. As shown in the two sets of experiments, only the MMA-UNet consistently provides the most accurate target information and exhibits strong capability in capturing pedestrian information hidden in the darkness in the distance. This indicates that the proposed asymmetric structure effectively preserves complementary information from different modal images. For other algorithms, segmentation failure occurs due to redundant information, such as capturing more detailed but weaker and more interfering pixel information from VI.


\subsection{Quantitative Analysis}

\textit{Fusion analysis.} \cref{tab1} lists the mean values of the five objective metrics in the two public test datasets. Overall, MMA-UNet achieves the best performance regardless of whether MSRS or M3FD is used. Specifically, the two metrics, SSIM and $Q_{cb}$, always exhibited the best value, indicating that MMA-UNet can retain the structure and contrast of the source image and generate fusion results that are the most consistent with the HVS. The rankings of VIF and PSNR fluctuate slightly but always exhibit high performance. The superior PSNR and VIF values demonstrate that our fusion results have strong robustness and high information fidelity. $Q_{abf}$ ranked third for both datasets, and MMA-UNet exhibited stable edge information preservation capabilities.

\textit{Detection analysis.}  \cref{tab2} presents the detection accuracy for all methods, including source images, on various categories in the M3FD. MMA-UNet demonstrates superior detection accuracy in both AP@0.5 and AP@[0.5:0.95]. Since various complex environments often affect the detection performance of single modalities, and MMIF could enhances detection robustness, fused images generally achieve better detection accuracy. Additionally, MMA-UNet exhibits superior detection accuracy across various categories, particularly for humans, indicating excellent preservation of detail and structural information for humans.

\textit{Segmentation analysis.}  \cref{tab3} shows the segmentation metrics for all methods, including source images, on various categories in the MSRS dataset. MMA-UNet achieves the best segmentation accuracy. Firstly, due to the superiority of MMIF, fusion methods generally achieve higher segmentation accuracy than single modal images. Additionally,  segmentation accuracy of MMA-UNet for each category ranks higher, indicating stable fusion performance and generation of rich semantic information in different scenarios.

\subsection{Ablation Analysis}
\cref{tab4} presents the objective metric mean obtained using the proposed method and various ablation strategies on the M3FD dataset. Overall, MMA-UNet achieved the best fusion performance. Specifically, after losing the guidance of the VI for IR-UNet feature extraction, both the symmetrical and asymmetrical fusion methods presented worse performance in preserving the structural information of the source images and fidelity than MMA-UNet. This is primarily attributed to the fact that guidance from the VI helps in aligning the information space distribution between the two modalities, which is beneficial for the subsequent fusion and fusion image reconstruction. Additionally, although E4 utilizes IR-UNet guided by VI, its excessively asymmetric approach discards too many shallow IR information, resulting in a significant decrease in the image fidelity, structural similarity, and edge information preservation. Finally, as evident from E5, interchanging asymmetric structures results in degraded model fusion performance. This can be attributed to significant disparities in the representation space for feature fusion, leading to a loss of details and semantic information in the fusion process.

The experiments above substantiate the effectiveness of the proposed asymmetric architecture, confirming the validity of the conclusion that VI feature extraction reaches deeper semantic space more rapidly than IR.

\begin{table}[h]
\caption{E1 represents MMA-UNet; E2 represents symmetrical fusion performed using IR-UNet without VI guidance; E3 represents asymmetric fusion performed using IR-UNet without VI guidance; E4 represents the asymmetric fusion of IR-UNet under VI guidance, where the first three layers of VI-UNet are fused with the last three layers of IR-UNet; E5 represents interchanging the asymmetric structure of E1.}
\centering
\begin{tabular}{c|lllll}
\hline
\multicolumn{1}{l|}{Experiment} & VIF            & SSIM           & PSNR            & $Q_{abf}$           & $Q_{cb}$            \\ \hline
\rowcolor{gray!20} E1                              & \textbf{0.430} & \textbf{0.764} & \textbf{16.335} & 0.626          & \textbf{0.529} \\
E2                              & 0.415          & 0.758          & 16.279          & \textbf{0.638} & 0.520          \\
E3                              & 0.419          & 0.755          & 16.313          & 0.617          & 0.517          \\
E4                              & 0.342          & 0.742          & 16.062          & 0.498          & 0.504          \\ 
E5                              & 0.344          & 0.757          & 16.247          & 0.449          & 0.498          \\ \hline
\end{tabular}
\label{tab4}
\end{table}

\section{Discussion}
\subsection{Conclusion}
In this study, we proposed an asymmetric UNet architecture for IVIF, providing a simple and effective method for fusing multi-modal image features. We analyze the differences in the spatial information distribution between the IR and VI modalities, and duduce the conclusion that there are discrepancy in the speed of extracting deep semantic features from different modalities within the same framework. We designed a cross-scale fusion rule based on different layer counts to address this issue. Subsequently, we designed a guidance mechanism for training IR-UNet and observed that simple guidance using VI features enhanced the efficiency of the deep semantic feature extraction in IR. The experimental results demonstrate that MMA-UNet outperforms the existing mainstream architectures, achieving good performance in MMIF and downstream tasks.

\subsection{Limitations and Future Work}
In this study, we revealed the difference in information space distribution between IR and VI and proposed a special, asymmetric, cross-scale fusion network architecture. However, the proposed asymmetric UNet requires manual analysis of the difference in information space distribution between different modal images, and designing the number of layers of feature fusion based on this difference. Therefore, in future work, in order to extend the proposed method to a broader range of visual tasks, we aim to devise an adaptive mechanism to selectively regulate the layer difference in the fusion of different modal features.

\section{Acknowledgments}
This research was supported in part by the National Natural Science Foundation of China under Grant 62201149, in part by the Joint Fund for Basic and Applied Basic Research of Guangdong Province under Grant 2023A1515140077, in part by the Natural Science Foundation of Guangdong Province under Grant 2024A1515011880, in part by the Guangdong Higher Education Innovation and Strengthening of Universities Project under Grant 2023KTSCX127, and in part by the Foshan Key Areas of Scientific and Technological Research Project under Grant  2120001008558, Guangdong, China.

\bibliographystyle{ACM-Reference-Format}
\bibliography{sample-base}

\appendix

\end{document}